# Small Object Detection: A Comprehensive Survey on Challenges, Techniques and Real-World Applications

Mahya Nikouei, Bita Baroutian, Shahabedin Nabavi, Fateme Taraghi, Atefe Aghaei, Ayoob Sajedi, Mohsen Ebrahimi Moghaddam[*]

Faculty of Computer Science and Engineering, Shahid Beheshti University, Tehran, Iran.

**Corresponding Author:** Mohsen Ebrahimi Moghaddam (PhD)

**Address:** Faculty of Computer Science and Engineering, Shahid Beheshti University, Tehran, Iran.

**Email:** m_moghadam@sbu.ac.ir

**Phone:** +98 912 140 5308

**Running Title:** A Survey on Small Object Detection

## Statements and Declarations

There are no conflicts of interests to declare.

**Abstract**

Small object detection (SOD) is a critical yet challenging task in computer vision, with applications like spanning surveillance, autonomous systems, medical imaging, and remote sensing. Unlike larger objects, small objects contain limited spatial and contextual information, making accurate detection difficult. Challenges such as low resolution, occlusion, background interference, and class imbalance further complicate the problem. This survey provides a comprehensive review of recent advancements in SOD using deep learning, focusing on articles published in Q1 journals during 2024-2025. We analyzed challenges, state-of-the-art techniques, datasets, evaluation metrics, and real-world applications. Recent advancements in deep learning have introduced innovative solutions, including multi-scale feature extraction, Super-Resolution (SR) techniques, attention mechanisms, and transformer-based architectures. Additionally, improvements in data augmentation, synthetic data generation, and transfer learning have addressed data scarcity and domain adaptation issues. Furthermore, emerging trends such as lightweight neural networks, knowledge distillation (KD), and self-supervised learning offer promising directions for improving detection efficiency, particularly in resource-constrained environments like Unmanned Aerial Vehicles (UAV)-based surveillance and edge computing. We also review widely used datasets, along with standard evaluation metrics such as mean Average Precision (mAP) and size-specific AP scores. The survey highlights real-world applications, including traffic monitoring, maritime surveillance, industrial defect detection, and precision agriculture. Finally, we discuss open research challenges and future directions, emphasizing the need for robust domain adaptation techniques, better feature fusion strategies, and real-time performance optimization. By consolidating recent findings and identifying research gaps, this survey serves as a valuable resource for researchers aiming to advance SOD methodologies.

# 1- Introduction

Small object detection(SOD) has emerged as a pivotal task in computer vision due to its significant role in various real-world scenarios. Unlike larger objects, small objects often contain limited spatial and contextual information, making their detection essential for applications where even minor elements can have a major impact. For instance, in safety-critical systems like autonomous vehicles, failing to detect small obstacles or pedestrians could result in catastrophic outcomes. The ability to accurately detect and localize small objects enhances the effectiveness and reliability of image-based decision-making systems, marking it as a key area of research (Cheng et al., 2023). Furthermore, advancements in this field contribute to the broader development of robust and scalable computer vision models that cater to diverse requirements in different domains.

The utility of SOD spans across a wide spectrum of applications, each critical to its respective field. In surveillance systems, accurately detecting small objects like unattended bags, individuals in crowds, or suspicious drones is crucial for maintaining public safety. In medical imaging, detecting small abnormalities such as tumors or microcalcifications significantly aids early diagnosis and treatment, potentially saving lives (Q. Feng et al., 2023). Similarly, in autonomous vehicles, recognizing small objects such as road signs, cyclists, or animals on the road ensures safe navigation. Moreover, in remote sensing, identifying small structures or vehicles in satellite imagery plays a vital role in urban planning and disaster response. These diverse applications underline the importance of developing highly accurate SOD methods tailored to meet domain-specific challenges.

Despite its importance, SOD remains a challenging task due to several inherent difficulties. The small size of the objects results in limited resolution and inadequate feature representation, making it difficult for models to differentiate them from the background. Moreover, small objects often appear in cluttered environments, where they may overlap with larger objects or be occluded by noise and artifacts, further complicating detection (Iqra et al., 2024). The class imbalance issue, where small objects are underrepresented in datasets, adds another layer of complexity, often leading to biased predictions. To address these challenges, researchers are exploring innovative techniques such as multi-scale feature extraction, SR approaches, and attention mechanisms to enhance detection accuracy for small objects in various contexts.

The primary objective of this survey is to provide a comprehensive overview of recent advancements in SOD using deep learning, focusing on articles published in Q1 journals (based on Scientific Journal Rankings (SJR)) during 2024 and 2025 indexed in Scopus with "Small Object Detection" in their titles. By reviewing these cutting-edge studies, we aim to analyze the challenges, techniques, datasets, evaluation metrics, and real-world applications related to the field. This survey not only consolidates existing knowledge but also identifies gaps and trends for future research. Compared to previous review articles (Cheng et al., 2023; Q. Feng et al., 2023; Iqra et al., 2024), our work offers a more detailed and updated perspective by integrating insights from the most recent publications. Additionally, our article highlights the superiority of the reviewed techniques in addressing specific challenges of SOD, such as low resolution, noise interference, and class imbalance, which have been less comprehensively covered in earlier surveys.

To structure the discussion effectively, the paper is organized as follows: Section 2 introduces the definitions and background of SOD, including its distinction from general object detection. Section 3 delves into the major challenges, including resolution limitations, background noise, small object size, and dataset imbalances. Section 4 provides an in-depth examination of advanced deep learning techniques and Recent trend, exploring innovations in neural network architectures, SR methods, data augmentation, transfer learning, and problem-specific approaches. Section 5 reviews key datasets like COCO, Pascal VOC, DOTA, and VisDrone, alongside evaluation metrics such as mAP, precision, recall, and F1-score, with an emphasis on their relevance to SOD. Section 6 outlines real-world applications, spanning areas like satellite imagery, video surveillance, medical imaging, and agriculture. Finally, Section 7 analyzes the future directions in this field. The paper concludes with a summary of findings and implications in Section 8. This comprehensive structure ensures that the survey not only serves as a reference for researchers but also facilitates the development of innovative solutions in SOD.

## 2- Definitions and Background

Small objects are typically characterized by their limited size, occupying a minimal number of pixels in an image, which makes their detection particularly challenging. The definition of small objects often depends on the dataset, application domain, and contextual criteria used in the analysis. Below, we explore various perspectives and definitions based on prominent datasets and research works.

### 2-1- Pixel-Based Definitions

A common approach to defining small objects is through their pixel dimensions or areas. For example, the MS COCO dataset, a widely-used benchmark in object detection, categorizes small objects as those with an absolute area smaller than 32×32 pixels, equivalent to less than 1,024 square pixels. This criterion is a standard reference point in many studies, as seen in the works utilizing MS COCO-based evaluations (S. Chen et al., 2024; Z. Chen, Ji, et al., 2024; Tian et al., 2024).

In the context of satellite and aerial imagery, small objects can appear even smaller. For instance, moving objects in satellite videos are often less than 20×20 pixels, which highlights the increased detection difficulty due to their diminutive size (S. Chen et al., 2024; Lei & Liu, 2024). Similarly, small objects in UAV imagery are frequently defined as those occupying a limited number of pixels (e.g., fewer than 32×32) and often consist of sparse, low-resolution features (Cao et al., 2024; R. Wang et al., 2024a).

### 2-2- Relative Size Criteria

Some definitions use the relative size of an object compared to the entire image area. For instance, objects occupying less than 1% of the image area are categorized as small (B. Liu & Jiang, 2024; L. Ni et al., 2024; Tong & Wu, 2024; Yang et al., 2024). This criterion becomes particularly relevant in high-resolution images where objects might have sufficient pixel counts but still appear small relative to the overall image dimensions.

Other studies define "small objects" using dynamic thresholds based on the dataset or application. For instance, in the TT-100K dataset, small objects are described as those smaller than 30×30 pixels in a 2048×2048 image, which corresponds to less than 1.5% of the total image resolution (Z. Zhu et al., 2024). This relative approach provides flexibility in adapting the definition to diverse imaging conditions.

### 2-3- Categorization of Small Objects

Several studies go beyond binary definitions to classify small objects into multiple subcategories:

- Tiny Objects: Defined as having areas up to 4500 square pixels.

- Small Objects: Larger than tiny objects but below specific thresholds based on datasets like MS COCO (Tong & Wu, 2024).
- Dense Small Objects (DSO): Overlapping objects in crowded scenes, such as urban areas captured by UAVs (C. Chen et al., 2024).

## 2-4- Challenges Highlighted by Definitions

While definitions provide a framework, many studies also emphasize the inherent challenges of SOD (F. Feng et al., 2024, 2025a; Shao et al., 2024):

- Low Pixel Count: Small objects often occupy fewer pixels, leading to limited feature information for recognition.
- Background Interference: The small scale of objects makes them susceptible to blending with complex backgrounds.
- Scale Variability: Small objects may vary in appearance across images due to changes in perspective or distance.

## 3- Challenges in SOD

Detecting small objects in images presents significant challenges across various fields, including industrial defect detection, surveillance, medical imaging, and remote sensing. These challenges arise due to a variety of factors, such as limited appearance information, occlusion, background interference, and the inherent limitations of existing detection models. This section explores the key challenges in SOD (Figure 1) and provides insights into current research and potential solutions.

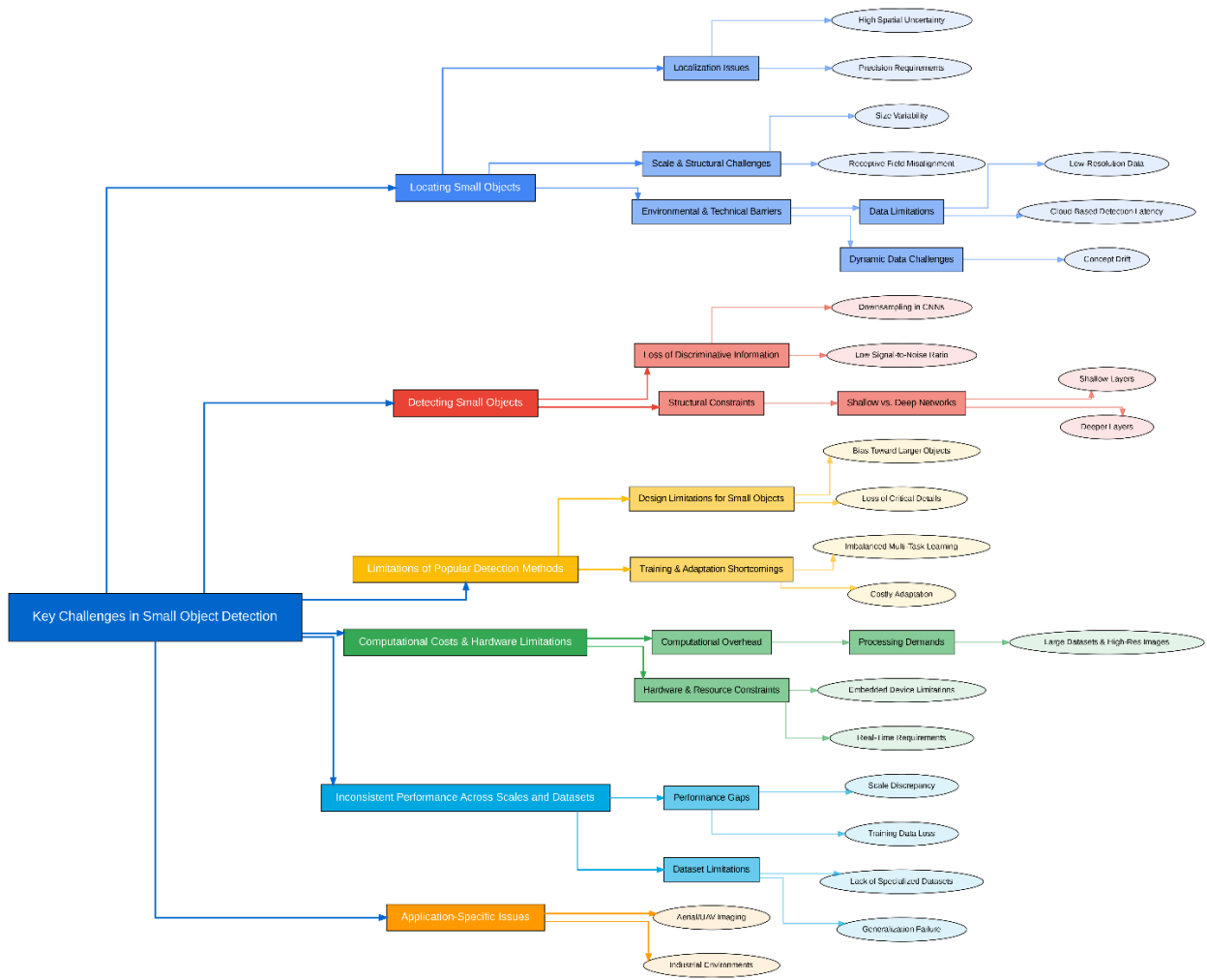

**Figure 1. Key Challenges in SOD and Related Issues**

### 3-1- Limited Appearance Information and Occlusion

Small objects often lack sufficient distinctive visual features that allow them to be differentiated from the background or other similar objects. This lack of identifiable characteristics, combined with frequent occlusion, makes detecting small objects difficult. For instance, in drone-captured images, the small targets may suffer from poor resolution, feature indistinguishability, and occlusion, which all significantly impair detection performance. Additionally, small objects tend to be more susceptible to various environmental factors, including lighting interference, blur, and occlusion. In scenarios such as coal mines, SOD is hindered by challenging environmental factors like low illumination, high dust, and water vapor density (Aibibu et al., 2023a; Jing, Zhang, Li, et al., 2024b; Ma et al., 2025; J. Wu et al., 2024).

Small objects often lack distinctive visual features, such as sharp edges, textures, colors, or patterns, which are crucial for differentiating them from the background or similar objects. Additionally, due to their size, they frequently suffer from poor resolution, making their features less discernible and harder to identify accurately. This lack of distinctive features and high resolution (Jing, Zhang, Li, et al., 2024b; Ma et al., 2025) leads to several challenges in detection.

One major issue is feature indistinguishability, where small objects become difficult to distinguish from the background or other objects, particularly in cluttered environments. They are also highly susceptible to occlusion (Ma et al., 2025; J. Wu et al., 2024), as they are frequently blocked by other objects, further complicating detection efforts. Moreover, small objects are more vulnerable to environmental factors such as lighting changes, motion blur, dust, low illumination (e.g., in coal mines), and water vapor, all of which degrade detection accuracy (Aibibu et al., 2023a).

These challenges significantly impair the performance of SOD systems in various applications. For instance, in drone imagery, issues like poor resolution and occlusion often result in missed or false detections. Similarly, in industrial settings such as coal mines, low light conditions and airborne particles exacerbate detection failures. Collectively, these factors hinder the reliability and effectiveness of SOD systems in real-world scenarios.

### 3-2- Challenges in Localization and Scale Variation

Locating small objects is inherently more difficult due to the larger number of possible locations and the need for higher localization accuracy. Small objects often face challenges in achieving high Intersection-over-Union (IoU) values with anchor boxes, making precise localization harder (Z. Zhou & Zhu, 2024). The small-scale nature of these objects requires that detectors exhibit an exceptionally fine-grained understanding of the image to accurately locate them. This is especially difficult when objects vary greatly in size and scale, leading to issues such as misalignment of the receptive field (Jing, Zhang, Li, et al., 2024b) or mismatched anchor boxes (S. Li et al., 2024). In environments like coal mines, small objects often lack high-resolution data, and detection methods that rely on centralized cloud computing experience significant latency. This is exacerbated by concept drift, where the characteristics of data change over time, which degrades the detection accuracy of small objects (J. Wu et al., 2024).

Locating small objects presents several key challenges, particularly in terms of localization difficulty and scale-related issues. One major challenge is the high spatial uncertainty, as the large number of possible locations for small objects increases the complexity of pinpointing them accurately (Z. Zhou & Zhu, 2024). Additionally, achieving precise localization is inherently difficult due to the small size of these objects. Detectors require higher localization accuracy, but obtaining high IoU values with anchor boxes is often challenging, leading to imprecise bounding box predictions (S. Li et al., 2024). Furthermore, detecting small objects demands a fine-grained understanding of images, requiring detectors to analyze details at an exceptionally high level. This level of detail is computationally intensive and adds to the complexity of the task (S. Chen et al., 2024; Sun et al., 2024).

Scale and structural issues further complicate the detection process. Small objects often exhibit significant variability in size and scale, which can cause mismatches between anchor boxes and the actual dimensions of the objects (D. Chen et al., 2024; Jing, Zhang, Li, et al., 2024b). This variability also leads to misalignment of the model's receptive field (the area "seen" by the detector) with the object's actual location or scale. Such misalignment reduces the reliability of localization, as the detector may fail to focus on the correct area or scale, impacting overall accuracy.

Environmental and technical problems also pose significant hurdles. In environments like coal mines, low-resolution data is a common issue, as small objects often lack high-quality imagery, further degrading detection performance (Song et al., 2024). Additionally, cloud-based detection methods, which rely on centralized computing, often suffer from latency, making real-time detection impractical. Dynamic data challenges, such as concept drift, further exacerbate the problem (J. Wu et al., 2024). Changes in data characteristics over time, such as shifts in dust levels, lighting conditions, or object appearance, can reduce detection accuracy. Models trained on older data may become outdated, struggling to adapt to these evolving conditions, which ultimately impacts their effectiveness in real-world scenarios.

### 3-3- Inefficiency in Feature Learning and Background Interference

The downsampling of input images in convolutional networks leads to a significant reduction in discriminative information, such as textures and edges. This reduction adversely affects the classification accuracy for small objects (R. Wang et al., 2024b),(S. Li et al., 2024). Additionally, small objects are often overshadowed by dense background interference, leading to a low Signal-to-Noise Ratio (SNR). In the case of industrial defect detection or UAV surveillance, small objects are frequently embedded in complex, noisy backgrounds, making them even harder to identify (D. Zhao et al., 2024). For example, in industrial settings, machinery defects may be surrounded by intricate visual data, while in aerial surveillance, small targets are frequently obscured by irrelevant background details. These complex backgrounds make it significantly harder to isolate small objects from the surrounding visual noise, reducing the effectiveness of detection systems in real-world applications (S. Chen et al., 2024; Zhang, Zhang, et al., 2024).

Structural limitations of detection networks, such as Feature Pyramid Networks (FPNs), also contribute to the challenge. These networks face a trade-off between shallow and deep layers. Shallow layers provide strong localization accuracy but lack rich semantic information, while deeper layers contain valuable semantic details but lose precise localization capabilities. This imbalance in performance leads to suboptimal detection (Yuan et al., 2019; Zheng et al., 2024), especially in environments where objects vary significantly in size, scale, or background complexity. As a result, detection networks struggle to achieve both accurate localization and meaningful semantic understanding simultaneously (F. Zhao et al., 2024).

### 3-4- Limitations of Popular Detection Methods

Existing object detection methods, such as those based on CNNs, were primarily designed for detecting larger objects and are thus not well-suited for SOD. These models often fail to preserve fine-grained details crucial for SOD. Popular detectors are also highly computationally intensive, particularly when working with high-resolution images, which significantly increases the resource demand and processing time (J. Liu et al., 2024).

Additionally, when training these models, there is often an imbalanced contribution between tasks in multi-task learning setups. This imbalance can hinder the model's ability to learn effective features for SOD, resulting in limited accuracy (S. Li et al., 2024). Furthermore, retraining existing methods to adapt them for SOD often requires significant modifications to the architecture, training process, or loss functions, which can be computationally expensive (C. Chen et al., 2024).

### 3-5- High Computational Costs and Hardware Resource Limitations

One of the biggest obstacles to SOD is the high computational overhead associated with processing large datasets, especially when high-resolution images are required. SOD in environments such as aerial imaging with UAVs introduces additional challenges, including limited computing resources on embedded devices and the need for efficient real-time analysis (Z. Chen, Ji, et al., 2024; Jing, Zhang, Liu, et al., 2024; J. Liu et al., 2024; Song et al., 2024). For instance, small-scale objects captured by drones often suffer from a lack of detailed information, and varying camera angles can lead to significant differences in object scale and density within a scene (Z. Chen, Ji, et al., 2024).

Moreover, embedded devices on unmanned surface vehicles (USVs) or in coal mines have limited computational power, making it difficult to deploy large, complex perception models. This is further complicated by the necessity for these systems to adapt to new object categories during missions, especially when dealing with limited samples (few-shot learning). Existing algorithms that perform well on SOD are often too resource-intensive and unsuitable for edge computing scenarios, where latency and real-time responsiveness are critical (Gao, Wang, et al., 2024; B. Liu & Jiang, 2024; R. Wang et al., 2024b; W. Wang et al., 2024).

### 3-6- Inconsistent Performance Across Different Scales and Datasets

Another significant challenge is the performance gap between small and large object detection. This gap becomes even more exacerbated when the training and testing datasets differ substantially in terms of object scale. Small objects often fall outside predefined anchor or grid regions, leading to a loss of training data. In addition, the continuous convolutional processes in advanced networks tend to cause feature disappearance, further hindering accurate detection (Tong & Wu, 2024; L. Zhou et al., 2024).

Existing methods often struggle when small objects are densely packed, as these objects may have limited or no distinguishing features in the image. This results in performance degradation when object scales vary dramatically between training and testing phases. The lack of specialized datasets focusing on small objects in large, complex areas, such as aerial images (Jiang et al., 2024) or large-scale industrial settings (Zou et al., 2024), exacerbates this issue.

### 3-7- Solutions and Emerging Approaches

In response to these challenges, several innovative approaches have been proposed. One promising solution is the introduction of motion-inspired cross-pattern learning, which enhances detection by incorporating both motion and visual cues to overcome the challenges of SOD in dynamic environments. Additionally, Multi-Granularity Detection (MgD) (D. Chen et al., 2024) frameworks have been introduced to address issues like poor detection performance and sample imbalance, allowing for more accurate and robust detection of small objects across different scales.

Furthermore, recent developments in neural networks, such as enhancing FPNs (Tian et al., 2024) to better fuse low-level and high-level features, offer a potential solution to the semantic gap that often affects SOD. These advancements aim to preserve crucial localization information while enriching semantic understanding, thus improving the accuracy of SOD.

## 4- Deep Learning Techniques for SOD

### 4-1- Recent Trends

Recent trends in Small Object Detection (SOD) focus on overcoming the challenges posed by complex environments through the development of advanced techniques aimed at improving model performance. As shown in Figure 2, these trends include feature extraction enhancement, feature fusion optimization, attention mechanisms, and advanced learning strategies. In the area of feature extraction, the use of optimized backbone architectures and the design of new feature enhancement modules have led to improved detection accuracy while reducing computational costs. Meanwhile, the effective fusion of features from multiple levels and scales—enabled by networks such as the Feature Pyramid Network (FPN) and lightweight integration modules—has contributed to better small object detection performance.

Attention mechanisms also play a crucial role in directing the model's focus toward key regions in the image. These mechanisms—ranging from channel and spatial attention to deformable and transformer-based attention—help suppress irrelevant background information, thereby enhancing the model's ability to detect small objects. Additionally, advanced learning strategies such as Knowledge Distillation (Nabavi, et al., 2024) and Reinforcement Learning have improved model adaptability in real-world deployment scenarios.

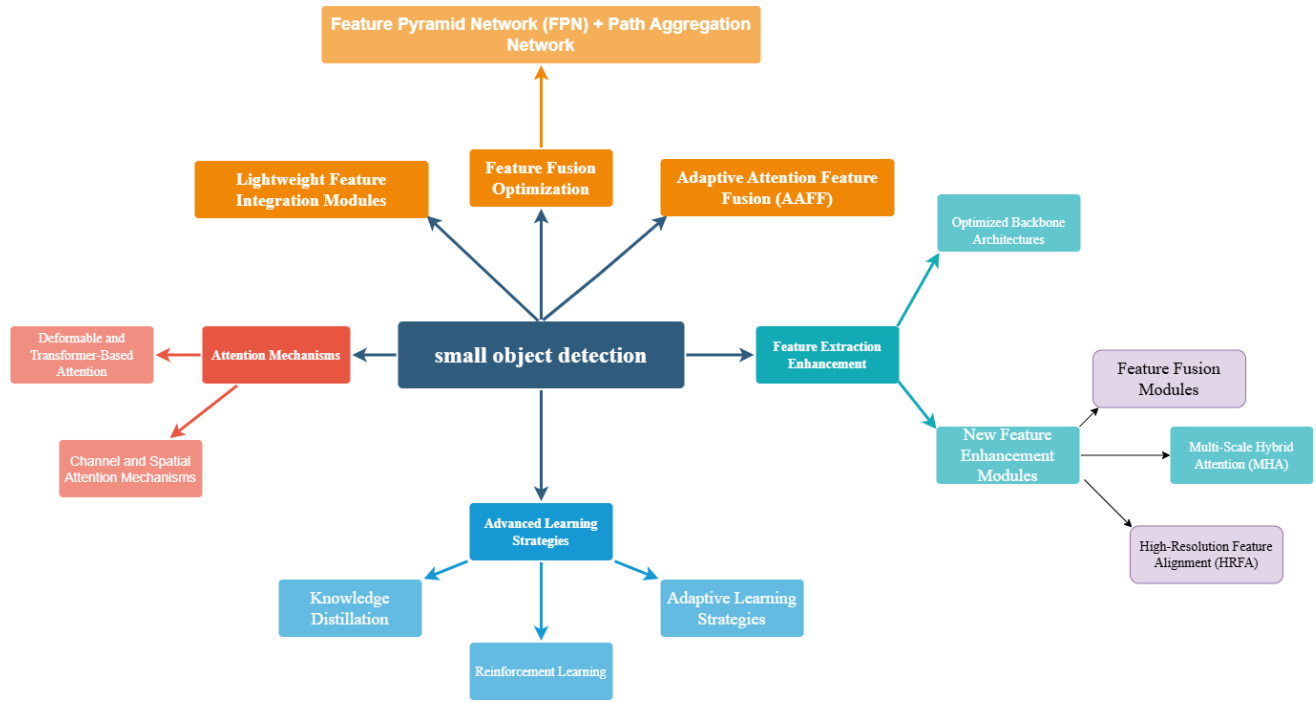

**Figure 2.** An overview of recent trends in SOD

Figure 3 presents the distribution of these emerging trends in recent research papers using a pie chart. According to the chart, the highest research focus has been on optimized backbone architectures, which account for 23.1% of the literature. This is followed by attention mechanisms at 18.5%, feature extraction enhancement at 16.9%, feature fusion optimization at 15.4%, advanced learning strategies at 13.8%, and multi-scale hybrid attention and high-resolution feature alignment at 12.3%.

A combined analysis of Figures 2 and 3 reveals that the primary focus of recent research in small object detection lies in strengthening core model architectures, improving feature fusion processes, and enhancing the model's attention to important image regions. These efforts reflect a unified strategy to improve the accuracy and efficiency of SOD models in multi-scale, noisy, and complex environments.

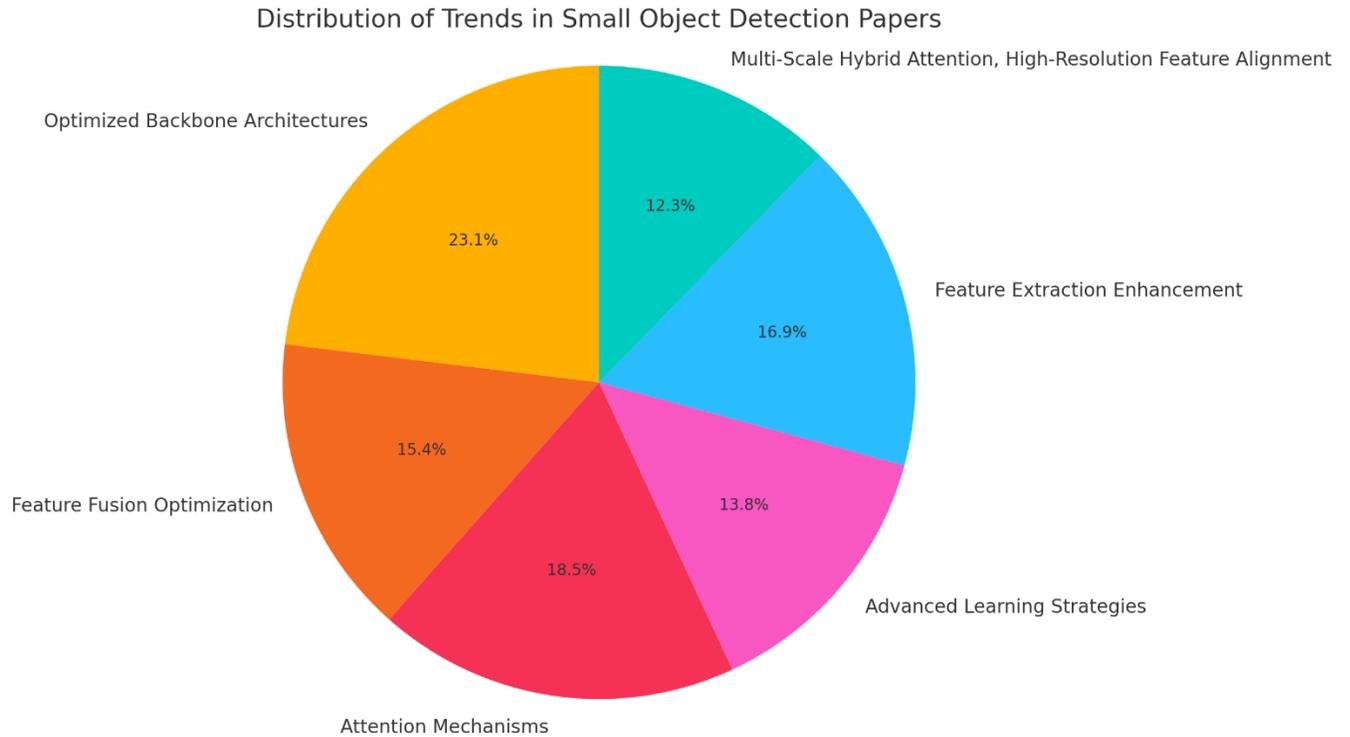

**Figure 3. Distribution of recent trends in SOD papers**

### 4-1-1- Model Optimization and Lightweight Architectures

Recent studies have focused on developing lightweight models tailored for real-time applications in resource-constrained environments. Approaches like modular design, parameter-sharing strategies, and compression techniques have enabled scalable, low-complexity architectures such as FFEDet (F. Zhao et al., 2024) and KDSMALL (W. Zhou et al., 2024). These models aim to balance computational efficiency with detection accuracy, ensuring deployment feasibility on drones, mobile devices, and edge computing platforms.

- Example: Integrating SR techniques to enhance image clarity while maintaining lightweight architectures.

### 4-1-2- Feature Fusion and Multi-Scale Detection

Advancements in multi-scale feature extraction and fusion techniques have significantly improved SOD performance. Methods like attention mechanisms, transformer-based detectors (e.g., DETR (DEtection TRansformer), Swin Transformer), and large kernel designs have enabled better contextual understanding and enhanced detection precision across scales (W. Zhou et al., 2024). Multi-frame feature fusion and cross-modal approaches (Jiang et al., 2024) (e.g., thermal and visible imaging fusion) have further strengthened small object detection in complex environments like haze, sandstorms, and maritime settings (Z. Chen, Shi, et al., 2024; W. Zhou et al., 2024).

- Example: Application of attention-driven fusion to UAV-captured scenarios for improved localization accuracy.

### 4-1-3- Anchor-Free and Transformer-Based Approaches

Anchor-free models have gained attention for their ability to simplify the detection process by directly predicting bounding boxes without predefined anchor points (S. Li et al., 2024). Transformer-based architectures, leveraging self-attention mechanisms, have also shown promise for SOD due to their ability to capture long-range dependencies and multi-scale features effectively . Hybrid CNN-transformer models are increasingly explored to harness the advantages of both frameworks (Z. Zhu et al., 2024).

- Example: Exploring DETR-like methods for post-processing to address the limitations of Non-Maximum Suppression (NMS).

### 4-1-4- Advanced Data Augmentation Techniques

To mitigate dataset imbalance and enhance generalization, advanced data augmentation methods such as (Generative Adversarial Networks)GAN-based synthetic image generation, hard augmentation strategies, and diffusion models have been widely adopted. These techniques enrich datasets, improve training robustness, and help address the scarcity of small-object-specific samples (Q. Liu et al., 2024).

- Example: Using GANs to generate synthetic UAV datasets for diverse environmental conditions.

### 4-1-5- Integration of Multi-Modal and Multi-Domain Data

SOD has increasingly leveraged multi-modal data (e.g., LiDAR (Light Detection and Ranging), radar, infrared, and visual images) to improve feature extraction and detection accuracy. Combining features across modalities enables the model to process richer representations and handle diverse scenarios effectively. This integration is particularly beneficial in real-world applications like industrial inspections and intelligent traffic systems (Jing, Zhang, Liu, et al., 2024).

- Example: Fusion of thermal and visible imaging features for industrial inspection tasks.

### 4-1-6- Knowledge Distillation and Self-Supervised Learning

KD has emerged as a key technique for improving the efficiency and real-time performance of SOD models. Additionally, self-supervised learning is being explored to reduce reliance on labeled data, enabling robust model training in unsupervised and semi-supervised scenarios (W. Zhou et al., 2024).

- Example: Deploying KD for lightweight SOD on UAVs.

## 4-2- Neural Network Architecture

Table 1 summarizes various papers and methods for SOD, focusing on strategies like feature extraction, feature fusion, and attention mechanisms. Each paper applies different techniques such as motion processing, deformable convolution, RL, and KD to improve SOD accuracy. These methods usually combine multiple approaches to enhance model performance under various conditions and reduce computational costs.

Table 1. Key Innovations in Recent Researches on SOD

| Research | Method | Recent trends Categorization | Technical justification |
|---|---|---|---|
| MICPL: Motion-Inspired Cross-Pattern Learning for Small-Object Detection in Satellite Videos (S. Chen et al., 2024) | Using the Motion Processing Module (MPM) to extract motion features and the Motion-Vision Alignment (MVA) Module to synchronize motion and visual information. | Optimized Backbone Architectures | The deployment of MPM for extracting motion features and improving the accuracy of small moving object detection. |
| | | Feature Fusion Optimization | the deployment of MVA for synchronizing visual and motion information and optimizing detection. |
| Object feedback and feature information retention for small object detection in intelligent transportation scenes (Tian et al., 2024) | "The use of SOPANet to preserve and fuse important features with residual connections and Small Object Intersection over Union (SOIoU) Loss to enhance localization accuracy." | Optimized Backbone Architectures | the deployment of SOPANet and Residual Connections to preserve important features. |
| | | High-Resolution Feature Alignment | the fusion of important feature information with Residual Connections to enhance accuracy. |
| | | Advanced Learning Strategies | the deployment of SOIoU Loss to improve small object localization. |
| ScorePillar: A Real-Time Small Object Detection Method Based on Pillar Scoring of LiDAR Measurement (Cao et al., 2024) | Utilizing a pillar-based encoding of LiDAR data combined with a Point-Score Feature Extraction (PSFE) module and PillarConv for efficient feature extraction, ScorePillar enhances real-time small object detection through attention-guided scoring, ResNet backbone, multiscale feature fusion, and atrous convolutions, with an anchor-based detection head for accurate localization. | Optimized Backbone Architectures | The model employs a ResNet-based backbone to efficiently extract both fine and coarse spatial features from sparse LiDAR data, which is crucial for capturing small object characteristics like pedestrians and bicycles. |
| | | Feature Extraction Enhancement | The LiDAR point cloud is encoded into vertical pillars, enabling 2D convolutional operations for higher computational efficiency. The PillarConv module further enhances feature representation by applying large-kernel convolutions, improving the detection of small, sparse objects. |
| | | Attention Mechanisms | Combines self-scoring (for each point based on spatial relevance) and mutual-scoring using a self-attention mechanism (ECANet), helping suppress noise and emphasizing key small object cues. |
| | | Feature Fusion Optimization | The neck module fuses features from multiple scales using atrous convolutions to expand the receptive field without losing resolution, which is critical for detecting small objects in sparse LiDAR data. |
| LA-YOLO: Bidirectional Adaptive Feature Fusion Approach for Small Object Detection of Insulator Self-Explosion Defects (B. Liu & Jiang, 2024) | Utilizing the Faster-C2f backbone network with Bidirectional Adaptive Feature Pyramid Network (Bi-AFPN-P2) for multi-scale feature fusion, combined with a DD-Head for spatial-aware convolutions and decoupling | Optimized Backbone Architectures | The backbone network Faster-C2f uses lightweight convolutional structures designed to minimize redundancy, enhancing the feature extraction efficiency, especially for small object detection in images. |
| | classification tasks to enhance small object detection for insulator self-explosion defects. | Feature Extraction Enhancement | The Faster-C2f backbone optimizes feature extraction with a reduction in parameters and FLOPs, focusing on fine-grained feature extraction for small objects like insulator self-explosion defects. |
| | | Feature Fusion Optimization | The Bi-AFPN-P2 neck component combines feature information across different scales and enhances fusion by applying bidirectional processing. This allows for effective detection of small objects in complex scenarios, such as insulator defects. |
| | | Adaptive Learning Strategies | The DD-Head uses spatial-aware convolutions to better localize small objects, improving the detection accuracy for insulator defects by decoupling classification tasks and reducing redundant computations. |
| An effective method for small objects detection based on MDFFAM and LKSPP (Shao et al., 2024) | "Utilizing a ResNet Backbone for multi-resolution feature extraction, a dual top-down pathway to enhance feature fusion across different levels, and a Feature-Aware Module (FAM) to reduce the semantic gap, along with deformable convolutions in the Head to improve the accuracy of detecting objects with varying shapes and sizes." | Optimized Backbone Architectures | the deployment of the ResNet Backbone for multi-resolution feature extraction. |
| | | Multi-Scale Hybrid Attention, High-Resolution Feature Alignment | the deployment of a dual top-down pathway for better multi-level feature fusion. |
| | | Attention Mechanism | the deployment of the FAM to reduce the semantic gap and focus on key features. |
| | | Advanced Learning Strategies | the deployment of deformable convolutions in the Head to enhance the accuracy of detecting small objects with varying shapes and sizes. |
| Feature aggregation network for small object detection (Jing, Zhang, Li, et al., 2024b) | "Utilizing Multi-Directional Feature Fusion Attention Mechanism (MDFFAM) to enhance spatial information and preserve spatial details, and Large Kernel Spatial Pyramid Pooling (LKSPP) to expand the receptive field and reduce computational complexity." | Optimized Backbone Architectures | the deployment of LKSPP to expand the receptive field and enhance feature extraction. |
| | | Multi-Scale Hybrid Attention, High-Resolution Feature Alignment | the deployment of MDFFAM for multi-directional feature fusion and spatial detail preservation. |
| | | Attention Mechanisms | the deployment of MDFFAM, a multi-directional spatial attention mechanism. |
| | | Advanced Learning Strategies | the deployment of LKSPP, which enhances the receptive field while maintaining computational efficiency. |
| A Lightweight Small Object Detection Method Based on Multilayer Coordination Federated Intelligence for Coal Mine IoVT (J. Wu et al., 2024) | "Utilizing lightweight models and federated networks for multi-layer coordinated learning and optimized SOD, aiming to reduce computational complexity and improve detection accuracy in various conditions." | Optimized Backbone Architectures | the deployment of lightweight models to reduce computational complexity and enhance feature extraction efficiency. |
| | | Feature Fusion Optimization | multi-layer coordination in federated learning for optimal feature fusion. |
| | | Advanced Learning Strategies | the deployment of federated learning to reduce complexity and improve detection accuracy in various conditions. |
| Libra-SOD Balanced label assignment for small object detection (Z. Zhou & Zhu, 2024) | Integrating spatial and temporal information through neighboring and distant signals, enhancing small object feature representation using Foreground Feature Alignment, Background Comparison, and Feature Amalgamation modules, and | Feature Extraction Enhancement | The deployment of Foreground Feature Alignment and Background Comparison |
| | | Feature Fusion Optimization | The deployment of Feature Amalgamation |
| | | Attention Mechanisms | The deployment of Channel and Spatial Attention |
| | reducing noise while improving detection accuracy in side-scan sonar images. | | |
| Toward High-Accuracy and Real-Time Two-Stage Small Object Detection on FPGA (S. Li et al., 2024) | "A two-stage architecture featuring low-resolution localization and high-resolution classification to improve accuracy and reduce computational cost." | Feature Extraction Enhancement | The deployment of a two-stage architecture for hierarchical feature processing |
| | | Feature Fusion Optimization | the fusion of low- and high-resolution information to enhance accuracy and reduce computational cost. |
| | | Advanced Learning Strategies | the deployment of an adaptive learning strategy where the first stage is more cost-efficient, and the second stage provides higher accuracy. |
| Multi-YOLOv8: An infrared moving small object detection model based on YOLOv8 for air vehicle (Sun et al., 2024) | "Utilizing a multi-input processing module to fuse current frames, suppress background, and extract motion and spatial features using optical flow, incorporating BiFormer for feature attention and GSConv for improved efficiency, enhancing localization accuracy with the α-WIoU v3 loss function, and increasing sensitivity to small objects with a dedicated detection layer." | Optimized Backbone Architectures | the deployment of a multi-input processing module for extracting spatial and motion features. |
| | | Multi-Scale Hybrid Attention, High-Resolution Feature Alignment | the fusion of current frames and the use of optical flow for improved spatiotemporal information processing. |
| | | Attention Mechanisms | the deployment of BiFormer, an advanced attention mechanism for processing key image features. |
| | | Adaptive Learning Strategies | the deployment of the α-WIoU v3 loss function and a dedicated detection layer to enhance the model's sensitivity to small objects. |
| Multi-granularity Detector for Enhanced Small Object Detection under Sample Imbalance (D. Chen et al., 2024) | Multi-level feature extraction through deformable convolution and sample balancing strategy using dynamic thresholds and layered decision-making. | Optimized Backbone Architectures | The deployment of Deformable Convolution |
| | | Attention Mechanisms | The deployment of Deformable Convolution |
| | | Adaptive Learning Strategies | the deployment of a sample balancing strategy with dynamic thresholds and layered decision-making |
| Small object detection in unmanned aerial vehicle images using multi-scale hybrid attention (Song et al., 2024) | "Utilizing a MHA structure to enhance SOD, incorporating MsA for cross-scale similarity, FEM for foreground feature enhancement, and depthwise separable channel attention (DSCA) for reducing channel redundancy, all integrated after the Neck." | Optimized Backbone Architectures | the deployment of MsA for extracting cross-scale similarities and optimizing small object features. |
| | | Multi-Scale Hybrid Attention, High-Resolution Feature Alignment | the deployment of FEM for enhancing foreground features and filtering out unnecessary information. |
| | | Attention Mechanisms | the deployment of MHA and DSCA, which play a key role in attention processing for important features. |
| Post-secondary classroom teaching quality evaluation using small object detection model (R. Wang et al., 2024b) | Optimized multi-scale feature extraction and fusion, utilization of attention mechanisms for accuracy enhancement, reduction of computational complexity for real-time efficiency, and simultaneous processing of spatial and temporal information. | Optimized Backbone Architectures | Optimized multi-scale feature extraction for efficient detection of small objects |
| | | Attention Mechanisms | The deployment of attention mechanisms to enhance object differentiation and mitigate background noise in SOD. |
| HV-YOLOv8 by HDPconv: Better lightweight detectors for small object detection (W. Wang et al., 2024) | "Replacing Halved Deep Pointwise Convolution (HDPConv) in the Backbone to enhance feature extraction and reduce complexity, and utilizing (View Group Shuffle Cross Stage Partial Network)VOV-GSCSP in the Neck for | Optimized Backbone Architectures | the deployment of HDPConv in the Backbone to enhance feature extraction and reduce complexity. |
| | | Multi-Scale Hybrid Attention, High-Resolution Feature Alignment | the deployment of VOV-GSCSP in the Neck for lightweight feature fusion and accuracy optimization. |
| | lightweight feature fusion to optimize accuracy and real-time performance." | Advanced Learning Strategies | Reduction in computational complexity and model optimization for real-time performance |
| MFFSODNet: Multiscale Feature Fusion Small Object Detection Network for UAV Aerial Images (Jiang et al., 2024) | Utilizing an optimized Multiscale Feature Fusion Small Object Detection Network(MFFSODNet) architecture for small-object detection in UAV images, incorporating Multiscale Feature Extraction Module (MSFEM) for rich multiscale feature extraction, Bidirectional Dense Feature Pyramid Network (BDFPN) for enhanced feature fusion, and a Modified Prediction Head for improved small object detection, all integrated into the YOLOv5-based framework for efficient and accurate performance. | Optimized Backbone Architectures | Due to the MSFEM, which uses multiple convolutional branches with different kernel sizes (e.g., 1×1, 3×3, 5×5), enhancing small object feature extraction and avoiding feature loss during downsampling. |
| | | Feature Extraction Enhancement | Due to the MSFEM, which captures fine-grained details and enables the network to extract multiscale object features effectively, improving detection performance, particularly for small objects. |
| | | Feature Fusion Optimization | Due to the BDFPN in the Neck, which enhances feature fusion by combining fine-grained shallow features with semantic deep features, facilitating the detection of small objects in complex scenarios with varying object scales. |
| KDSMALL: A lightweight small object detection algorithm based on knowledge distillation (W. Zhou et al., 2024) | "Utilizing EfficientNet in the Backbone to enhance feature extraction with multi-scale features, CBAM in the Neck to apply spatial and channel attention and improve feature fusion, and KD in the Head to transfer knowledge from a larger model to a smaller one for increased accuracy and maintained efficiency." | Optimized Backbone Architectures | the deployment of EfficientNet in the Backbone for optimized multi-scale feature extraction. |
| | | Multi-Scale Hybrid Attention, High-Resolution Feature Alignment | the deployment of CBAM in the Neck for optimal feature fusion and the application of channel and spatial attention. |
| | | Attention Mechanisms | the deployment of CBAM, which implements a spatial and channel attention mechanism. |
| | | Advanced Learning Strategies | the deployment of KD in the Head to enhance accuracy and reduce computational complexity. |
| A Small-Object Detection Based Scheme for Multiplexed Frequency Hopping Recognition in Complex Electromagnetic Interference (Z. Chen, Shi, et al., 2024) | Utilizing an optimized Receptive Field Refinement Module (RFRM)-CenterNet architecture for small-object detection, incorporating ResNet50 in the Backbone for feature extraction, RFRM for enhancing multi-scale features, and a specialized Head with three sub-networks for precise frequency-hopping signal detection, all integrated for complex spectrogram analysis. | Optimized Backbone Architectures | Due to the use of ResNet50 as the backbone for extracting primary features from spectrograms, optimized for small object detection in frequency-hopping signals. |
| | | Feature Extraction Enhancement | Due to the Receptive RFRM in the Neck, utilizing convolutional layers with varying dilation rates to enhance multi-scale features and improve the detection of small objects. |
| | | Feature Fusion Optimization | Due to the RFRM's ability to integrate multi-scale features effectively, enhancing the representation of small objects in spectrograms. |
| | | Optimized Backbone Architectures | Due to the use of ResNet50 as the backbone for extracting primary features from spectrograms, optimized for small object detection in frequency-hopping signals. |
| MAE-YOLOv8-based small object detection of green crisp plum in real complex orchard | "Leveraging Efficient Multi-scale Attention (EMA) for better object-background distinction, | Optimized Backbone Architectures | the deployment of AFPN for preserving low-level features and enhancing feature extraction. |
| environments (Q. Liu et al., 2024) | AFPN for preserving low-level features, and minimum point distance intersection over union (MPDIoU) Loss for improved localization in occlusion scenarios." | Multi-Scale Hybrid Attention (MHA), High-Resolution Feature Alignment | the fusion of multi-scale features with AFPN to retain important information. |
| | | Attention Mechanisms | the deployment of EMA for better object-background separation. |
| | | Advanced Learning Strategies | the deployment of MPDIoU Loss to enhance localization accuracy in occlusion scenarios. |
| ESOD: Efficient Small Object Detection on High-Resolution Images (K. Liu et al., 2025) | Optimized multi-scale feature extraction and fusion, utilization of attention mechanisms for accuracy enhancement, reduction of computational complexity for real-time efficiency, and simultaneous processing of spatial and temporal information. | Optimized Backbone Architectures | Optimized multi-scale feature extraction for efficient detection of small objects |
| | | Attention Mechanisms | The deployment of attention mechanisms to enhance object differentiation and mitigate background noise in SOD. |
| TA-YOLO: a lightweight small object detectionmodel based on multi-dimensional trans-attentionmodule for remote sensing images (M. Li et al., 2024) | Utilizing multi-head channel and spatial trans-attention (MCSTA) to extract attention features across multiple dimensions (channel and spatial), integrating features in the Neck via PAN, and optimizing performance. | Optimized Backbone Architectures | the deployment of MCSTA, a Transformer-based attention module for multi-dimensional feature extraction. |
| | | Multi-Scale Hybrid Attention, High-Resolution Feature Alignment | the fusion of features using PAN, which facilitates information aggregation in the Neck. |
| | | Attention Mechanisms | the deployment of MCSTA, which integrates both channel and spatial attention. |
| Adaptive Feature Fusion and Improved Attention Mechanism-Based Small Object Detection for UAV Target Tracking (Xiong et al., 2024) | "Utilizing Adaptive Feature Fusion Mechanism (AFFM) to balance features of small and large objects, Soft Pooling to minimize information loss during feature extraction, and Subspace Attention to enhance spatial localization of small objects while suppressing background noise." | Optimized Backbone Architectures | the deployment of Soft Pooling to minimize information loss during feature extraction. |
| | | Multi-Scale Hybrid Attention, High-Resolution Feature Alignment | the deployment of AFFM for optimal feature fusion of small and large objects. |
| | | Attention Mechanisms | the deployment of Subspace Attention to enhance spatial localization and suppress background noise. |
| An adaptive lightweight small object detection method for incremental few-shot scenarios of unmanned surface vehicles (B. Wang et al., 2024) | "Utilizing a MHA structure to enhance SOD, including MsA for cross-scale similarity, Feature Extraction Module(FEM) for foreground feature enhancement, and DSCA for reducing channel redundancy, all integrated after the Neck." | Optimized Backbone Architectures | the deployment of MsA for extracting cross-scale similarities and optimizing small object features. |
| | | Multi-Scale Hybrid Attention, High-Resolution Feature Alignment | the deployment of FEM for enhancing foreground features and filtering out unnecessary information. |
| | | Attention Mechanisms | the deployment of MHA and DSCA, which play a key role in attention processing for important features. |
| A Small-Object Detection Model Based on Improved YOLOv8s for UAV Image Scenarios (J. Ni et al., 2024) | Utilizing an optimized backbone architecture based on YOLOv8s, incorporating Scale Compensation Feature Pyramid Network (SCFPN) for multi-scale feature fusion, and introducing an ultra-SOD layer (P2) for improved small object detection, all integrated after the Neck. | Optimized Backbone Architectures | Due to the use of Convolutional and Bottleneck layers. |
| | | Feature Extraction Enhancement | Due to the use of SCFPN |
| | | Feature Fusion Optimization | Due to the use of SCFPN's weighted feature fusion. |
| | | Adaptive Learning Strategies | Due to the use of the ultra-small-object detection layer (P2) in the Head. |
| A Heatmap-Supplemented R-CNN Trained Using an Inflated IoU for Small Object Detection(Butler & Leung, 2024) | Utilizing a dual-backbone architecture combining a conventional R-CNN convolutional backbone and an hourglass network for enhanced | Optimized Backbone Architectures | Due to the integration of a conventional R-CNN backbone alongside an hourglass backbone for enhanced small object proposal generation. |
| | small object region proposal generation, incorporating objectness heatmaps and an inflated IoU training strategy to refine small object localization, all integrated after the Neck. | Feature Extraction Enhancement | Due to the use of objectness heatmaps in the parallel hourglass backbone to refine object localization. |
| | | Feature Fusion Optimization | Due to the modification of FPN to better capture small object features across multiple scales. |
| | | Adaptive Learning Strategies | Due to the Inflated IoU Training Strategy, which applies a Gaussian-based multiplier to artificially enhance IoU values for small objects, improving training stability and detection accuracy. |
| YOLOv8-QSD: An Improved Small Object Detection Algorithm for Autonomous Vehicles Based on YOLOv8 (H. Wang et al., 2024) | Utilizing the enhanced YOLOv8 architecture as a base, incorporating Diverse Branch Block (DBB) for scalable feature extraction, C2f-DBB for efficient small object detection, Bidirectional Feature Pyramid Network (BiFPN) for improved feature fusion, Q-block with Query Mechanism for accurate localization, and DyHead for optimized multi-scale attention, all integrated into the YOLOv8 framework for robust small object detection. | Optimized Backbone Architectures | Enhanced YOLOv8 Framework: The backbone architecture builds on the original YOLOv8 structure, incorporating C2f modules, decoupled detection heads, and an anchor-free structure, optimizing small object detection for autonomous vehicles. |
| | | Feature Extraction Enhancement | Due to the DBB, which uses a multi-branch structure during training for better scalability and a single-branch structure during inference, balancing complexity and performance to efficiently detect small objects. |
| | | Feature Fusion Optimization | Due to the BiFPN, which enhances feature fusion across scales, improving the representation of small objects across multiple levels in the network. |
| | | Optimized Backbone Architectures | Enhanced YOLOv8 Framework: The backbone architecture builds on the original YOLOv8 structure, incorporating C2f modules, decoupled detection heads, and an anchor-free structure, optimizing small object detection for autonomous vehicles. |
| SONet: A Small Object Detection Network for Power Line Inspection Based on YOLOv8 (Shi et al., 2024) | Utilizing YOLOv8 architecture with the Multi-Branch Dilated Convolution Module (MDCM) for multi-scale feature extraction, Adaptive Attention Feature Fusion (AAFF) for enhanced feature fusion using attention mechanisms, and the novel β-CIoU loss function for optimized bounding box regression, improving small object detection for power line inspection. | Optimized Backbone Architectures | The SONet model is built upon the YOLOv8 framework, with core advancements like MDCM for multi-scale feature extraction. This ensures improved performance, particularly in power line inspection where small object detection is critical. |
| | | Feature Extraction Enhancement | The MDCM is employed in SONet to capture multi-scale features by using dilated convolutions with multiple branches. This helps detect small objects that might be spatially spread across multiple scales, ensuring better recognition of details in complex power line inspection images. |
| | | Feature Fusion Optimization | The AAFF module uses attention mechanisms to refine and fuse features across different scales. This adaptive approach enhances the model's ability to focus on critical regions of interest, such as small defects on power lines, improving small object detection. |

### 4-3- Clarity and Visual Information Improvement

Improving the clarity and visual information in computer vision tasks is crucial for enhancing the performance of models, particularly in areas such as object detection, segmentation, and image classification. One of the key objectives in addressing these challenges is to ensure that models can extract and utilize fine-grained details from images, which is especially important when dealing with small or low-resolution objects. This section explores three main techniques for improving clarity and visual information: SR for image quality enhancement, the use of multi-scale information, and the fusion of information from different layers of neural networks (D. Chen et al., 2024).

### 4-3-1- SR for Image Quality Enhancement

SR is a technique used to improve the resolution of an image, effectively increasing the clarity of fine details that might otherwise be lost in low-resolution images. SR aims to recover high-resolution details from a series of low-resolution images, typically utilizing deep learning-based models such as CNNs (K. Liu et al., 2025). These models learn to upscale the low-resolution input into a higher-resolution output, using the contextual information in the image to fill in missing details.

In the context of SOD, SR plays a critical role in improving the quality of images where tiny objects might be indistinguishable due to resolution constraints. By enhancing the resolution of the input image, SR allows models to discern minute details and achieve more accurate localization and classification of small objects. Techniques such as the use of GANs for SR have also gained popularity, as GAN-based approaches can generate realistic high-resolution images by learning the underlying distribution of the image data (Y. Zhao et al., 2024).

Moreover, SR methods are often coupled with other image enhancement techniques, such as denoising and deblurring, to ensure that the super-resolved images do not introduce artificial artifacts that could degrade the overall model performance (Jobaer et al., 2025).

### 4-3-2- Utilizing Multi-Scale Information

Multi-scale information refers to the ability to extract and process features at various scales from an image, ensuring that both fine-grained details and broader context are captured. Many computer vision tasks, including SOD, benefit from multi-scale feature extraction because objects in images can appear at different sizes, and detecting them effectively requires information from multiple levels of abstraction (D. Liao et al., 2025a; Y. Zhao et al., 2024).

Models that incorporate multi-scale information typically use networks that process images at different resolutions or feature pyramid structures that merge multi-level feature maps (F. Feng et al., 2025b; Song et al., 2024; B. Wang et al., 2024). This allows the network to recognize both small and large objects by detecting patterns across different scales. For example, in SOD, fine-grained details at a low resolution (captured by early layers in the network) can be combined with more abstract, high-level features (captured by deeper layers) to better detect small objects.

One of the key challenges when using multi-scale information is balancing the spatial and semantic information at each scale. Fine-grained details can sometimes get lost when transitioning to higher resolutions, and models must be designed to efficiently handle this information flow. Feature pyramids, such as FPN, are commonly employed to address this issue by allowing models to combine high-resolution fine-grained features with coarser contextual information from deeper layers (Z. Chen, Ji, et al., 2024; Gao, Li, et al., 2024; Q. Liu et al., 2024).

### 4-3-3- Fusion of Information Across Network Layers

In modern deep learning models, the fusion of information from different layers within the network is a crucial technique for improving performance. The feature maps generated by various layers in a neural network capture different aspects of the input image, from low-level edges and textures to high-level semantic information (Zhang, Zhang, et al., 2024; Z. Zhou & Zhu, 2024). Fusing these feature maps allows the network to leverage a more comprehensive understanding of the image, which is essential for tasks such as SOD, where small objects may only be represented in early layers of the network.

Several strategies for layer information fusion exist, including concatenation, addition, and attention mechanisms. Attention mechanisms (Jiang et al., 2024; Zhang, Zhang, et al., 2024; F. Zhao et al., 2024), in particular, have shown promise in dynamically weighting the importance of features from different layers. By assigning more weight to crucial layers or regions in the image, attention mechanisms can help models focus on areas that contribute the most to the task at hand. For example, attention modules like the Spatial Attention Module (SAM) or Channel Attention Module (CAM) enable models to emphasize spatial regions or channels that are most relevant for the detection of small objects.

Another approach to information fusion is the use of Feature Fusion Networks (FFNs) (Xie et al., 2024), which combine the outputs from different layers or feature maps to create a more holistic representation of the image. By capturing both high-level semantic and low-level spatial information, these networks can improve clarity and help models detect objects more effectively, particularly when those objects are small or poorly represented in individual layers.

### 4-4- Data Augmentation and Synthetic Data

### 4-4-1- Data Augmentation Techniques for SOD

Data augmentation is a widely used strategy to artificially expand the training dataset by applying a variety of transformations to existing images. In the context of SOD, specific augmentation techniques are employed to simulate real-world scenarios and help the model focus on identifying small-scale features (Q. Liu et al., 2024).

- Rotation and Scaling: Small objects often appear in varying orientations and sizes. Applying random rotations and scaling adjustments helps models generalize better to different spatial representations of small objects.

- Cropping and Zooming: Focused cropping and zooming augmentations ensure that small objects remain within the image frame while forcing the model to learn fine-grained details.

- Padding and Contextual Cropping: Augmentations like padding can introduce a buffer around small objects, while contextual cropping emphasizes objects and their surroundings to improve contextual awareness.

- Brightness, Contrast, and Noise Adjustments: Adjusting brightness and contrast or adding Gaussian noise ensures that models can handle variations in lighting and image quality, which are common in real-world conditions.

- Random Erasing and Occlusion Simulation: Simulating occlusions by partially erasing regions of an image teaches the model to detect small objects even when partially obscured.

Advanced augmentation methods, such as MixUp, CutMix, and Mosaic, have been particularly impactful for SOD (Jing, Zhang, Li, et al., 2024b; J. Wu et al., 2024; Zhang, Zhang, et al., 2024). These techniques combine multiple images into a single composite image, creating complex scenarios that enhance the model's ability to discern small objects within cluttered backgrounds.

**4-4-2- Synthetic Data Generation for SOD**

While data augmentation transforms existing data, synthetic data generation creates entirely new training samples, addressing the issue of limited datasets for SOD. This approach is especially beneficial when capturing real-world images of small objects is impractical or expensive (Q. Liu et al., 2024).

- GANs: GANs have been extensively used to create high-quality synthetic images with small objects. By training a generator to produce realistic small objects and their backgrounds, GAN-based approaches can augment datasets with rare or diverse object instances (D. Chen et al., 2024; Jing, Zhang, Li, et al., 2024b; Tian et al., 2024).

- Computer-Generated Imagery (CGI): CGI-based synthetic data allows for complete control over the positioning, size, and texture of small objects in images. This flexibility makes it possible to generate datasets tailored specifically for SOD tasks (Zhang, Zhang, et al., 2024).

- Simulators and 3D Rendering: Simulation platforms such as Unity, CARLA, or Blender enable researchers to generate complex environments populated with small objects. These tools allow for adjustments in lighting, camera angles, and occlusions, ensuring that synthetic datasets closely mimic real-world conditions (Tian et al., 2024).

Synthetic datasets (Q. Liu et al., 2024) also provide pixel-perfect annotations for bounding boxes, segmentation masks, and keypoints, which are essential for training detection models. This automated labeling significantly reduces the cost and time associated with manual annotation processes.

### 4-4-3- Advantages of Data Augmentation and Synthetic Data in SOD

The use of data augmentation and synthetic data generation provides several benefits for detecting small objects:

- Enhanced Generalization: Models trained on augmented and synthetic datasets are better equipped to generalize to unseen environments, especially when small objects appear in challenging contexts (Q. Liu et al., 2024).

- Improved Robustness: Exposure to variations in scale, occlusions, and lighting improves the robustness of detection algorithms (Q. Liu et al., 2024; F. Zhao et al., 2024).

- Addressing Data Scarcity: Synthetic data enables the creation of large datasets in domains where real-world data collection is limited, such as aerial imagery or medical imaging (Q. Liu et al., 2024; F. Zhao et al., 2024).

- Cost Efficiency: Generating synthetic data is often more cost-effective than collecting and annotating large-scale real-world datasets (Q. Liu et al., 2024; F. Zhao et al., 2024).

- Class Imbalance Mitigation: Augmentation and synthetic data generation can be tailored to oversample underrepresented small object classes, ensuring a balanced dataset (F. Zhao et al., 2024).

### 4-4-4- Challenges and Research Directions

Despite their benefits, these techniques face certain challenges:

- Domain Gap: A significant difference in visual characteristics between synthetic and real-world images can reduce model performance when applied to real-world data. Bridging this gap through domain adaptation techniques is an active area of research (S. Chen et al., 2024).

- Over-Augmentation: Excessive or unrealistic augmentations may introduce noise or unrealistic scenarios, negatively impacting model training. Careful tuning of augmentation parameters is crucial (S. Chen et al., 2024; Liang et al., 2020).

- Scalability of Synthetic Data: Creating high-quality synthetic datasets with diverse small objects requires significant computational resources and expertise (S. Chen et al., 2024).

Future research directions include exploring automated data augmentation pipelines (e.g., AutoAugment), improving synthetic data realism through advanced generative models (e.g., diffusion models), and incorporating multi-modal synthetic data to enhance contextual learning for SOD.

### 4-5- Multi-task and Transfer Learning

### 4-5-1- Multi-Task Learning for SOD

Multi-task learning focuses on training a model to perform multiple related tasks simultaneously, enabling the model to share knowledge and features across tasks. For SOD, this approach helps the model learn complementary information from auxiliary tasks, ultimately improving detection performance. Key multi-task learning strategies are:

- Shared Feature Representations: By sharing the backbone of a neural network across tasks such as object detection, segmentation, and classification, the model can learn more robust and generalized features that benefit SOD (M. Li et al., 2024).

- Auxiliary Tasks: Tasks like SR, edge detection, and depth estimation are often integrated with SOD to enhance the model's ability to focus on fine-grained details (L. Zhou et al., 2024). For instance:
    - SR as an Auxiliary Task: Generating high-resolution versions of low-resolution input images can improve the visibility of small objects, allowing the model to extract more discriminative features.
    - Edge Detection: Learning to identify edges or boundaries can guide the model to locate small objects more precisely.

- Multi-Scale Learning: Incorporating tasks at different scales, such as coarse-scale detection for context and fine-scale detection for details, helps improve the detection of small objects.

- Task-Specific Heads: Separate output heads for each task allow the model to focus on task-specific objectives while sharing a unified backbone for feature extraction.

Multi-task learning offers significant advantages , particularly for SOD, by leveraging information from multiple tasks to develop enriched feature representations. This enhances the model's ability to detect subtle and small objects while also acting as a form of regularization to mitigate overfitting through task-related data. Furthermore, it improves computational efficiency by enabling a single model to handle multiple tasks, reducing the resource demands associated with training separate models. However, multi-task learning also presents challenges that must be addressed to ensure optimal performance. Task conflicts may arise when different tasks have competing objectives, potentially leading to suboptimal performance, necessitating effective task balancing strategies such as dynamic loss weighting (S. Li et al., 2024). Additionally, designing appropriate architectures and training strategies becomes increasingly complex when handling diverse tasks, requiring careful model design and optimization techniques.

**4-5-2- Transfer Learning for SOD**

Transfer learning involves reusing a model trained on a large-scale dataset for a new, often smaller, target dataset. This approach is particularly beneficial for SOD, where labeled data is often scarce. Pre-trained models for transfer learning include:

- Feature Transfer: Models pre-trained on large datasets such as ImageNet or COCO can provide robust feature extractors that are fine-tuned for SOD tasks. The early layers of such models capture generic features like edges and textures, which are transferable across tasks (Butler & Leung, 2024; Y. Li, Yang, et al., 2024).

- Fine-Tuning: Transfer learning typically involves freezing the lower layers of the pre-trained model and fine-tuning the higher layers on the target dataset to specialize in detecting small objects (Tian et al., 2024; W. Zhou et al., 2024).

- Domain-Specific Pre-Training: For certain applications, such as aerial imagery (Jing, Zhang, Li, et al., 2024a) or medical imaging (Zou et al., 2024), pre-training on a domain-specific dataset further enhances the transferability of features.

Techniques in transfer learning are:

- KD: This technique involves transferring knowledge from a large, complex teacher model to a smaller, lightweight student model. KD is particularly effective for SOD when computational resources are limited (Jobaer et al., 2025; W. Zhou et al., 2024).

- Cross-Domain Transfer: Models trained on datasets with large objects can be adapted to detect small objects by leveraging techniques like domain adaptation and domain-invariant feature extraction (Cao et al., 2024).

- Few-Shot Learning: Transfer learning can be combined with few-shot learning techniques to improve performance in scenarios where only a few labeled examples of small objects are available (B. Wang et al., 2024).

Transfer learning offers several benefit (Aibibu et al., 2023b; Butler & Leung, 2024), particularly in the context of SOD. It significantly reduces the amount of labeled data required for training, enhancing data efficiency. Additionally, models initialized with pre-trained weights converge faster and demonstrate improved performance compared to models trained from scratch. Transfer learning also improves generalization by leveraging knowledge from large-scale datasets, allowing the model to perform better in unseen scenarios. However, transfer learning comes with its own set of challenges. One major issue is domain shift, where a mismatch between the source and target domains can undermine the effectiveness of transfer learning. Techniques like domain adaptation (Nabavi et al., 2023; Nabavi, et al., 2024) are necessary to address this problem. Additionally, fine-tuning on small target datasets may lead to overfitting, particularly when the dataset does not adequately represent the variability of the task.

### 4-5-3- Integration of Multi-Task and Transfer Learning

Combining multi-task learning with transfer learning can amplify the benefits of both approaches. For example, pre-trained multi-task models can be fine-tuned on a specific SOD task, leveraging both shared knowledge across tasks and pre-trained feature representations. This integration provides a powerful framework for handling the challenges of SOD.

The integration of multi-task and transfer learning has demonstrated significant promise in various applications. In aerial imagery, it has proven effective in detecting small vehicles, animals, and buildings from satellite or drone images. In medical imaging, it enhances the identification of small lesions, tumors, or abnormalities in medical scans (C. Chen et al., 2024; W. Zhou et al., 2024). Additionally, in surveillance, it facilitates the detection of small objects such as weapons or unattended bags in crowded scenes (Cao et al., 2024; Z. Chen, Shi, et al., 2024; D. Liao et al., 2025b).

Future research directions include the development of automated task balancing algorithms to dynamically address conflicting objectives in multi-task learning (S. Li et al., 2024). Another key area is domain adaptation, where improving techniques to handle domain shifts in transfer learning will be crucial (Cao et al., 2024). Additionally, exploring the integration of few-shot learning with multi-task learning for SOD presents a promising avenue (Yuan et al., 2019). Lastly, research into continual transfer learning aims to build models capable of adapting to new tasks and domains while retaining knowledge of previously learned tasks.

## 5- Datasets and Evaluation Metrics

### 5-1- Datasets

This table (Table 2) summarizes various datasets used for SOD. Each dataset addresses challenges like size variation, occlusions, and complex backgrounds, and is intended for tasks such as surveillance, autonomous driving, and real-time tracking. These datasets, captured from UAVs and satellites, offer valuable data for training models to handle real-world detection challenges.

Table 2. Summary of SOD Datasets

| Dataset | Categories | Challenges | Size | Purpose |
|---|---|---|---|---|
| VisDrone (P. Zhu et al., 2018) | 10 object categories, including pedestrians, bicycles, vehicles, and buses | Varying spatial density, occlusion, environmental factors (weather, lighting), and complex urban and rural backgrounds. | 10,209 images in total with 6,471 for training, 548 for validation, and 3,190 for testing. | Aerial dataset for SOD captured by UAVs, addressing challenges like object size variation, occlusion, and complex backgrounds, with real-world factors such as weather and lighting changes. |
| DIOR (K. Li et al., 2020) | 20 object categories, including vehicles, pedestrians, and other common objects in urban and rural environments. | Aerial object detection with varying sizes, occlusions, and complex backgrounds. | Large-scale dataset with over 1 million instances. | Focused on object detection in aerial images, particularly for UAV-based applications, offering a diverse set of challenges for robust model training. |
| DOTA (Ding et al., 2022) | 15 object categories, including vehicles, buildings, ships, and planes. | Large-scale aerial images with varying object sizes, complex backgrounds, and occlusions. | 2,806 high-resolution aerial images with 188,282 object instances. | Aerial dataset with 15 object categories, featuring large-scale images and various challenges like size variation, occlusions, and complex backgrounds. Suitable for training object detection models in satellite and UAV imagery. |
| VEDAI (Razakarivony & Jurie, 2016) | Primarily focused on vehicle detection, including various vehicle types like cars, trucks, and buses. | Object detection in aerial imagery with varying vehicle sizes, occlusions, and complex backgrounds. | 1,268 high-resolution aerial images with 2,950 vehicle instances. | Designed for vehicle detection in UAV and satellite imagery, providing valuable data for training detection models in diverse real-world environments. |
| TinyPerson (X. Yu et al., 2020) | Focuses on pedestrian detection, specifically targeting small-sized individuals in complex environments. | Detecting tiny, occluded, and distant pedestrians in crowded and cluttered settings. | 72,000 images with annotations for small pedestrian detection. | Designed to address the challenges of detecting small persons in crowded and occluded scenes, useful for applications like surveillance and crowd monitoring. |

| Dataset | Object Categories | Challenges | Size | Purpose |
|---|---|---|---|---|
| MS COCO (Lin et al., 2014) | 80 object categories, including people, animals, vehicles, and everyday objects. | Detecting objects in complex scenes with varying sizes, occlusions, and background clutter. | Over 330,000 images, with more than 2.5 million object instances annotated. | Designed for object detection, segmentation, and captioning tasks, widely used for training and evaluating models in real-world image analysis. |
| SODA (C. Chen et al., 2017) | Focuses on SOD, with annotations for various object types such as vehicles, pedestrians, and others. | Designed for detecting small objects in images, with challenges like occlusion, size variation, and complex backgrounds. | Includes thousands of images with annotated small objects, categorized into subsets based on object size (e.g., extremely small, relatively small). | Aimed at improving SOD, particularly in UAV and aerial imagery, where objects often occupy only a few pixels. |
| DroneCrowd (Wen et al., 2021) | Focuses on crowd detection and pedestrian tracking in urban environments. | Detecting and counting small, densely packed crowds with occlusions in aerial imagery. | Contains 13,416 annotated objects across several high-resolution aerial images. | Designed for crowd counting and SOD in aerial views from UAVs, useful for surveillance, security, and crowd management applications. |
| UAVDT (H. Yu et al., 2020) | Includes various object categories such as vehicles, pedestrians, and cyclists. | Object detection and multi-object tracking in aerial images, with challenges like occlusions, scale variations, and complex backgrounds. | 50,000 frames across various sequences. | Focuses on object detection and tracking for UAV-captured videos, aimed at developing models for real-time applications like surveillance and traffic monitoring. |
| BDD100K (F. Yu et al., 2018) | The dataset includes 10 object categories, such as cars, pedestrians, bicycles, traffic signs, traffic lights, and other driving-related objects. | The challenges in BDD100K include various weather conditions (rain, snow, sunny), lighting variations, occlusion, and the presence of moving objects. Additionally, environmental complexities like urban streets and rural roads add to the challenges. | The dataset consists of 100,000 images captured from cameras mounted on vehicles in various driving conditions. | The purpose of BDD100K is to support research in autonomous driving and object detection in real-world driving scenarios. It is used for training deep learning models in object recognition, traffic sign detection, and simulating various driving conditions. |
| WSODD (Z. Zhou et al., 2021) | The dataset includes various object categories that might appear on the water surface, such as boats, logs, debris, and other objects floating on water. | Key challenges in WSODD include variations in object size, occlusion due to other objects, reflections on the water surface, and complex environmental conditions like changing water levels, lighting, and weather. | The dataset typically includes a collection of images captured by UAVs, often with annotations of the objects on the water surface. The size of the dataset can vary depending on the source, but it usually contains a large number of images with varying object sizes and conditions. | The purpose of WSODD is to support research in detecting small or floating objects on water surfaces in aerial imagery. This is useful for applications such as monitoring waterways, environmental protection, and detecting hazards or obstacles on water surfaces using UAVs. |
| AI-TOD (J. Wang et al., 2020) | AI-TOD typically involves detecting small objects across a variety of categories such as vehicles, animals, pedestrians, and other objects that appear tiny due to their size or distance in the image. | The main challenges in AI-TOD include detecting objects that occupy very few pixels in images, dealing with occlusions, variations in scale, low resolution, and complex backgrounds. Additionally, tiny objects are often difficult to distinguish from noise, making accurate detection harder. | The dataset or approach in AI-TOD usually involves high-resolution images or videos where the objects of interest are much smaller than the average objects in traditional object detection tasks. The datasets used for AI-TOD are often large-scale, involving thousands of images with annotations for tiny objects. | The purpose of AI-TOD is to develop and refine algorithms that can accurately detect and classify very small objects, even in challenging conditions. Applications of AI-TOD include surveillance, remote sensing, medical imaging, and autonomous driving, where detecting tiny objects (e.g., pedestrians, animals, or road hazards) is crucial for safety and decision-making. |
| KITTI (Y. Liao et al., 2021) | Includes cars, pedestrians, cyclists, and other objects commonly found in urban and rural environments. | Object detection, stereo vision, and tracking in driving scenarios with varying lighting, weather, and motion. | 40,000 images, with 20,000 for training and 4,000 for testing. | Primarily used for autonomous driving research, providing 3D object annotations and a variety of real-world challenges for detection and tracking tasks. |

### 5-1-1- Aerial Image Datasets

Aerial imagery datasets are extensively used for SOD due to their real-world relevance and challenging features, including high variability in object sizes, occlusion, and dense clustering.

#### 5-1-1-1- VisDrone

The VisDrone dataset is one of the most widely adopted benchmarks for SOD in aerial imagery (P. Zhu et al., 2018). Collected by the Machine Learning and Data Mining Lab at Tianjin University, it provides:

- Diversity: 10 object categories, including pedestrians, bicycles, vehicles, and buses, with varying levels of spatial density and occlusion.
- Scale: Over 10,209 images divided into training (6,471), validation (548), and testing (3,190) subsets.
- Challenges: Captures diverse weather conditions, lighting variations, and complex urban and rural backgrounds.
- HFYSOJD

#### 5-1-1-2- DIOR

The DIOR (Dataset for Object Detection in Optical Remote Sensing Images) dataset (K. Li et al., 2020) focuses on remote sensing tasks and offers:

- Scale: 23,463 images and 192,472 annotated instances across 20 object categories.
- Variety: Images are collected from multiple sensors under different conditions, making it suitable for testing detection robustness.

#### 5-1-1-3- DOTA

The DOTA (Dataset for Object Detection in Aerial Images) dataset (Ding et al., 2022) is another key resource:

- Multiple Versions: Includes DOTA-v1.0, v1.5, and v2.0, each with incremental improvements in sample size and annotations.
- Features: Over 15 object categories, such as small vehicles, ships, and storage tanks, annotated with rotated bounding boxes for precise localization.

#### 5-1-1-4- VEDAI

The Vehicle Detection in Aerial Imagery (VEDAI) dataset (Razakarivony & Jurie, 2016) is designed for small vehicle detection in aerial images:

- Focus: Emphasizes multi-class detection of vehicles under varying resolutions.
- Challenges: Small object sizes and cluttered environments.

### 5-1-2- UAV-Based Datasets

UAVs provide rich datasets for SOD due to their ability to capture images from various altitudes and angles.

### 5-1-2-1- UAVDT

The Unmanned Aerial Vehicle Detection and Tracking (UAVDT) dataset (H. Yu et al., 2020) focuses on moving object detection:

- Scale: Includes over 100 videos with annotations for vehicles under different scenarios like highways and parking lots.
- Challenges: Motion blur, small object sizes, and varying weather conditions.

### 5-1-2-2- SODA-D

The SOD Dataset (SODA-D) (C. Chen et al., 2017) is specifically designed for detecting small objects captured by UAVs:

- Features: Includes diverse weather and lighting conditions to mimic real-world scenarios.
- Object Categories: Nine object classes such as pedestrians, cars, and bicycles, annotated with object size categories (large, medium, small, very small).

### 5-1-2-3- DroneCrowd

This dataset (Wen et al., 2021) targets crowd tracking tasks:

- Applications: Useful for detecting and tracking small human objects in crowded urban areas.
- Characteristics: High spatial density and occlusion levels make it challenging.

### 5-1-3- Autonomous Driving Datasets

Datasets for autonomous driving often include small object annotations for pedestrians, road signs, and vehicles.

### 5-1-3-1- BDD100K

The Berkeley DeepDrive dataset (F. Yu et al., 2018) contains 100,000 diverse driving scenes:

- Focus: Multi-class detection for objects like traffic lights, pedestrians, and cars.
- Challenges: Varying weather, lighting, and urban/rural settings.

### 5-1-3-2- KITTI

KITTI (Y. Liao et al., 2021) provides annotated datasets for autonomous driving:

- Features: Focuses on 3D object detection but includes small object categories like pedestrians and bicycles.

### 5-1-4- Specialized Datasets for SOD

#### 5-1-4-1- TinyPerson

TinyPerson (X. Yu et al., 2020) is specifically designed for detecting small human objects:

- Scale: Contains images with high-resolution annotations for small pedestrians.
- Applications: Useful for surveillance and crowd analysis.

#### 5-1-4-2- WSODD

The Water Surface Object Detection Dataset (WSODD) (Z. Zhou et al., 2021) targets objects in maritime environments:

- Focus: Includes small objects like boats and buoys, often occluded by waves or environmental factors.

#### 5-1-4-3- AI-TOD

The AI-TOD (Artificial Intelligence for Tiny Object Detection) dataset (J. Wang et al., 2020) includes:

- Diversity: Real-world scenarios with annotations for very small objects in aerial imagery.
- Applications: Designed for detecting objects like vehicles and ships in challenging environments.

### 5-1-5- Synthetic Datasets

Synthetic datasets (Z. Chen, Shi, et al., 2024) address the scarcity of labeled data for SOD by simulating realistic conditions.

#### 5-1-5-1- Motion Blur Synthetic Dataset

This dataset introduces motion blur to simulate UAV image capture conditions:

- Features: Covers varying levels of motion blur, enabling models to generalize better for blurred real-world images (H. Wu et al., 2024).

#### 5-1-5-2- Self-Collected Datasets

Several studies generate synthetic datasets (Zou et al., 2024) for specific tasks, such as detecting industrial components, insulators, or maritime objects:

- Examples: Includes datasets for power line insulators, copper wires, and steel bars, annotated under controlled experimental conditions.

### 5-1-6- Comparative Insights and Challenges

The reviewed datasets highlight key challenges in SOD:

- Low Resolution: Small objects often occupy only a few pixels, making feature extraction difficult (Zhang, Ye, et al., 2024).

- Occlusion and Clutter: Dense spatial arrangements hinder accurate detection (H. Wang et al., 2024).

- Diverse Conditions: Variability in weather, lighting, and object orientation adds complexity (Jiang et al., 2024).

- Domain-Specific Needs: Different applications require tailored datasets, such as aerial imagery, driving scenarios, or industrial monitoring.

### 5-1-7- Future Directions

For future directions in SOD, expanding datasets to cover underrepresented domains is essential for enhancing model robustness. Developing more diverse and large-scale datasets can improve generalization across various scenarios. Additionally, advancements in synthetic data generation, particularly high-fidelity simulations, can supplement real-world datasets, addressing issues of data scarcity. The integration of multi-modal data, such as thermal imaging and LiDAR, offers complementary information that enhances detection accuracy beyond conventional RGB images (Z. Chen, Shi, et al., 2024; Xu et al., 2024). Furthermore, adopting dynamic annotation techniques, including polygon masks and 3D bounding boxes, can improve the precision of small object labeling, leading to better detection performance in complex environments.

## 5-2- Evaluation Metrics

Evaluating the performance of SOD systems requires specialized metrics to account for the unique challenges posed by small and tiny objects, such as their low resolution, limited features, and high susceptibility to occlusion. This section reviews the most commonly used evaluation metrics in the field of SOD, emphasizing their utility and adaptability across different datasets and scenarios.

### 5-2-1- Metrics Commonly Used Across General Object Detection

- **Average Precision (AP):**
  The AP metric measures the area under the Precision-Recall curve for individual classes. It is widely adopted due to its ability to quantify precision across different recall levels. Specific thresholds like $AP50_{\{50\}}50$, $AP75_{\{75\}}75$, and $AP50:95_{\{50:95\}}50:95$ (a COCO-style metric) are often employed, where the subscript indicates the Intersection over Union (IoU) threshold.

    - AP5(50): IoU = 0.50

    - AP(75): IoU = 0.75

- AP(50:95): Average AP across IoUs ranging from 0.50 to 0.95 at intervals of 0.05.

- **Mean Average Precision (mAP):**
  The mAP aggregates AP values across all classes and is the primary metric for comparing model performance in multi-class object detection. It is particularly useful for datasets with varying object sizes. Metrics such as mAP(50), mAP(75), and mAP(50:95) are frequently reported.

### 5-2-2- Metrics for Size-Specific Detection

To address the challenges of detecting objects of varying sizes, especially small and tiny ones, metrics are tailored to measure performance across object scales.

- **APS_SS, APM_MM, APL_LL:**
  These metrics, used in COCO and similar datasets, evaluate performance for objects categorized by size (C. Chen et al., 2024):

  - APS_SS: For small objects ($area < (32 \times 32)\ Pixels$).
  - APM_MM: For medium-sized objects ($area\ between\ (32 \times 32)\ and\ (96 \times 96)\ Pixels$).
  - APL_LL: For large objects ($area > (96 \times 96)\ Pixels$).

- **APT_TT:**
  Introduced for datasets like TinyPerson, APT_TT measures **AP** specifically for tiny objects, which are even smaller than those categorized as "small" in COCO (Jing, Zhang, Li, et al., 2024b; K. Liu et al., 2025).

### 5-2-3- Tracking-Specific Metrics

For SOD in tracking tasks, specialized metrics are used (S. Chen et al., 2024):

- T-AP10, T-AP15, T-AP20:
  These metrics evaluate tracking performance at IoU thresholds of 10%, 15%, and 20%, emphasizing the continuity and accuracy of tracking small, often indistinct objects.

- T-mAP:
  Similar to mAP but adapted for tracking scenarios, this metric accounts for the temporal consistency of detected objects across frames.

### 5-2-4- Core Metrics for Precision and Recall

- Precision, Recall, and F1-Score:
  These metrics are foundational in SOD, providing insights into a model's ability to minimize false positives (Precision), identify all relevant instances (Recall), and balance the trade-off between the two (F1-Score) (Q. Liu et al., 2024).

### 5-2-5- Specialized Metrics for Small Object Datasets

Certain datasets and tasks define additional metrics tailored to their characteristics:

- Processed Pixel Number Percentage (PPN):
  This metric, relevant to applications like aerial imagery, measures the proportion of processed pixels during detection, emphasizing computational efficiency (Fang et al., 2024).

- SNR and Coefficient of Variation (CV):
  Metrics such as SNR and CV evaluate the quality of the detection process, particularly in noisy or low-resolution datasets (Zou et al., 2024).

### 5-2-6- Dataset-Specific Metric Implementation

- COCO and PASCAL VOC:
  These datasets employ AP, AP(50), AP(75), and size-specific metrics (APS_SS, APM_MM, APL_LL).

- VisDrone and TinyPerson (Z. Chen, Ji, et al., 2024; Jing, Zhang, Li, et al., 2024b; K. Liu et al., 2025):
  Metrics such as APT_TT, mAP, and APS_SS are essential for evaluating models on these datasets due to the predominance of small and tiny objects.

- DIOR and DOTA:
  These remote sensing datasets utilize mAP, AP(50), AP(75), and APS_SS to assess object detection across complex and diverse scenes (Gao, Wang, et al., 2024; Shi et al., 2024; Z. Zhou & Zhu, 2024).

- Custom Metrics for Specialized Datasets:
  Some studies develop metrics tailored to their specific datasets, such as datasets with artificially generated motion blur or varying object sizes. For instance, metrics such as Mean Average Recall (mAR) and image quality-based metrics are used for challenging datasets like SODA-D (Y. Li, Li, et al., 2024; H. Wang et al., 2024; Z. Zhou & Zhu, 2024) or WSODD.

### 5-2-7- Computational Efficiency Metrics

- Frames Per Second (FPS):
  FPS measures the real-time capability of a detection system, often reported alongside AP or mAP to highlight trade-offs between accuracy and speed (K. Liu et al., 2025).

- Inference Time:
  The average time required to process an image or a frame, often provided to showcase the practicality of the detection system for resource-constrained applications (Fang et al., 2024).

## 6- Applications and Real-world Use Cases

SOD has emerged as a critical area of research and development, addressing challenges in detecting objects with minimal size and fine detail in diverse scenarios. Its applications span across numerous domains, leveraging advancements in machine learning and computer vision to solve complex real-world problems. Below is a comprehensive review of the applications and use cases of SOD.

### 6-1- Remote Sensing and Aerial Surveillance

One of the most prominent applications of SOD is in remote sensing, where high-resolution satellite and aerial imagery enables large-scale analysis (Y. Li, Yang, et al., 2024; Zheng et al., 2024). In urban monitoring and planning, SOD helps identify vehicles, infrastructure components, and buildings, aiding in efficient urban development and management. It also plays a critical role in disaster management by detecting damage, such as collapsed buildings or blocked roads, to facilitate rapid response during natural disasters. In wildlife protection, UAVs equipped with SOD capabilities assist in monitoring animal populations and preventing poaching in sanctuaries. Additionally, environmental monitoring benefits from tracking changes in vegetation, water bodies, and other natural resources. In military reconnaissance, satellite and UAV imagery can identify enemy assets, including small vehicles or aircraft, enhancing situational awareness. Furthermore, maritime surveillance relies on SOD to track ships, vessels, and other marine objects, supporting ocean investigations and security patrols.

### 6-2- UAV and Drone Applications

Unmanned Aerial Vehicles (UAVs) have become widely utilized in SOD due to their ability to capture high-resolution imagery from varying altitudes. In surveillance and security, UAVs enable real-time monitoring for border patrol, crowd management, and public safety (S. Chen et al., 2024; W. Zhao et al., 2024). They also play a crucial role in forest fire detection, identifying small-scale flames and hotspots for early fire warning systems. In traffic monitoring, UAVs assist in tracking vehicles, bicycles, and pedestrians to analyze traffic flow, manage parking lot utilization, and address congestion issues (Jing, Zhang, Liu, et al., 2024). Additionally, UAV-based search and rescue operations help locate missing persons or vehicles in disaster zones or dense environments, improving emergency response efficiency (C. Chen et al., 2024). In agriculture, UAVs contribute to monitoring crop health, detecting pest infestations, and assessing irrigation systems, ultimately optimizing agricultural productivity.

### 6-3- Autonomous Systems

SOD is essential in autonomous driving, robotics, and unmanned systems, as it ensures safe navigation and operation. In autonomous driving, detecting distant or small-scale traffic signs, pedestrians, bicycles, and obstacles is vital for road safety (J. Liu et al., 2024). For autonomous navigation, SOD aids USVs and autonomous drones in path planning and obstacle avoidance within complex environments (B. Wang et al., 2024). Additionally, in Simultaneous Localization and Mapping (SLAM), detecting small objects improves the accuracy of mapping and navigation, enhancing the capabilities of robotic systems in dynamic settings.

### 6-4- Industrial Applications

In industrial settings, SOD plays a critical role in ensuring the quality and safety of manufacturing processes and products. It is used in defect detection, where subtle surface defects on mechanical equipment, PCBs, and vehicle components are identified to maintain product integrity (Xu et al., 2024). Additionally, SOD aids in aerospace inspection, helping to detect imperfections in aircraft structures and components. In cable manufacturing, this technology is utilized to monitor the production process for defects or inconsistencies, ensuring high-quality outputs.

### 6-5- Surveillance and Security

SOD significantly enhances security systems by identifying subtle threats or anomalies in real-time. In video surveillance, it helps detect intruders, suspicious objects, or abnormal activities in high-resolution CCTV footage . For border security, SOD is employed to monitor small-scale movements or objects across borders using drones or stationary sensors, improving surveillance efficiency and response times (F. Zhao et al., 2024).

### 6-6- Environmental and Natural Resource Monitoring

SOD plays a significant role in sustainable environmental management and resource utilization. In natural resource surveys, it helps identify and monitor small-scale changes in resources, such as mineral deposits or water levels. For geological mapping, SOD is used to detect subtle geological features, aiding in exploration and analysis (F. Feng et al., 2024). In forestry management, it contributes to tracking tree health, identifying illegal logging activities, and monitoring wildlife movement, ensuring better conservation and resource management.

### 6-7- Medical Imaging and Biological Analysis

In healthcare and biological research, SOD is crucial for identifying minute patterns or anomalies. In medical imaging, it is used to detect early signs of diseases, such as small tumors or strokes, in diagnostic imagery, aiding in timely diagnosis and treatment (Zou et al., 2024). For biological analysis, SOD helps monitor biological tissues and cellular structures, supporting research and enhancing diagnostic accuracy.

### 6-8- Maritime and Oceanographic Applications

The marine domain extensively benefits from SOD for monitoring and navigation in challenging environments. In autonomous ship navigation, SOD is used to identify objects like buoys, debris, or other vessels, ensuring safe navigation. For ocean investigation, it aids in tracking marine life, underwater features, and small-scale oceanographic changes, providing valuable insights into the marine ecosystem. Additionally, in security patrols, SOD is employed to monitor coastal and offshore areas, helping to detect illegal activities or potential threats, enhancing maritime security (Shao et al., 2024).

**6-9- Intelligent Transportation Systems**

SOD plays a crucial role in ensuring the efficiency and safety of transportation systems. In the context of traffic sign detection, it enables the recognition of small traffic signs under various weather and lighting conditions, which is essential for the effective operation of autonomous vehicles (Jiang et al., 2024; H. Wang et al., 2024). Furthermore, pedestrian flow monitoring utilizes SOD to estimate pedestrian density and movement, aiding in urban traffic management and enhancing public safety (D. Liao et al., 2025b). Additionally, railway and roadway inspections benefit from this technology by detecting minor faults or obstacles on tracks and roads, thereby contributing to the maintenance and safety of transportation infrastructure.

**6-10- Specialized Applications in Agriculture and Forestry**

SOD plays a pivotal role in precision agriculture and forestry management by enabling targeted monitoring. In smart agriculture, it aids in the identification of pests, small weeds, or crop diseases in real-time, allowing for timely interventions to optimize crop yield and health. Similarly, in forest monitoring, SOD is used to detect illegal logging activities, track wildlife movement, and monitor changes in forest health from aerial imagery, supporting effective conservation and management efforts (Zheng et al., 2024).

**7- Future Directions**

Recent advancements in SOD have focused on several key areas to enhance performance and deployment feasibility across diverse platforms and environments (Table 3). One area of development is lightweight and scalable networks, where efforts are directed at optimizing architectures to reduce computational complexity and improve parameter-sharing mechanisms, alongside developing lightweight backbone networks specifically tailored for SOD (D. Liao et al., 2025b; J. Liu et al., 2024). Additionally, feature representation has been enhanced through multi-scale feature fusion techniques and the incorporation of advanced modules like dynamic feature focusing networks, improving the detection of small objects in cluttered scenes. Leveraging SR models has also shown promise in extracting fine-grained features in low-resolution or degraded environments (García-Murillo et al., 2023; Xiao et al., 2024). In terms of transformer-based innovations, the integration of transformers into SOD models addresses limitations of CNN-based architectures, particularly in multi-scale feature processing, and combining CNN and transformer frameworks offers a balance of efficiency and precision (Jiang et al., 2024; L. Zhou et al., 2024).

Further advancements are seen in data augmentation and domain adaptation, where generative techniques like GANs and diffusion models are used to create realistic small-object datasets under varying environmental conditions, while domain adaptation techniques ensure robust detection performance across unseen datasets and challenging scenarios (Cao et al., 2024; Jobaer et al., 2025; Q. Liu et al., 2024). Advanced label assignment and loss functions are also a focus, with efforts to optimize label assignment methods for imbalanced datasets and small object classes, alongside introducing sophisticated loss functions to improve small-object localization and reduce false detection rates (Z. Zhou & Zhu, 2024).

To address the detection of small, fast-moving objects, cross-pattern and temporal context integration is being explored. This includes integrating motion pattern mining and temporal context from video sequences, as well as developing algorithms that incorporate temporal information for improved performance in dynamic environments (B. Zhao et al., 2024). Furthermore, applications in emerging domains are being expanded, with research exploring new fields such as medical imaging, ocean surveillance, and precision agriculture, as well as incorporating SOD models into intelligent systems like SLAM for improved localization accuracy in real-world applications (J. Ni et al., 2024; F. Zhao et al., 2024). Finally, real-time performance optimization focuses on minimizing computational overhead through pruning, quantization, and architectural refinements, aiming for lightweight deployment on resource-constrained devices such as UAVs, industrial robots, and IoT sensors (Jiang et al., 2024; J. Ni et al., 2024; H. Wang et al., 2024; F. Zhao et al., 2024).

**Table 3. Future Directions in SOD**

| Direction | Details |
|---|---|
| Lightweight and Scalable Networks (D. Liao et al., 2025b; J. Liu et al., 2024) | -Enhancing network efficiency by further optimizing lightweight architectures, reducing computational complexity, and improving parameter-sharing mechanisms.<br>- Developing lightweight backbone networks for SOD. |
| Enhancing Feature Representation (F. Feng et al., 2024; Xiao et al., 2024) | - Improving multi-scale feature fusion techniques and incorporating dynamic feature focusing networks for better small object representation in complex scenes.<br>- Leveraging SR models for low-resolution environments. |
| Transformer-Based Innovations (Jiang et al., 2024; L. Zhou et al., 2024) | - Exploring transformer integration to address CNN-based limitations and enhance multi-scale feature processing.<br>- Combining CNN and transformer frameworks to balance efficiency and precision. |
| Data Augmentation and Domain Adaptation (Cao et al., 2024; Jobaer et al., 2025; Q. Liu et al., 2024) | - Expanding generative techniques like GANs and diffusion models for realistic small-object datasets under varying conditions.<br>- Investigating domain adaptation for robust performance across unseen datasets. |
| Advanced Label Assignment and Loss Functions (Z. Zhou & Zhu, 2024) | - Optimizing label assignment for imbalanced datasets and small object classes.<br>- Introducing sophisticated loss functions to improve small-object localization and reduce false detection rates. |
| Cross-Pattern and Temporal Context Integration (B. Zhao et al., 2024) | - Integrating motion pattern mining and temporal context from video sequences to enhance detection of small, fast-moving objects.<br>- Developing algorithms that incorporate temporal information for robust performance. |

| Applications in Emerging Domains (J. Ni et al., 2024; F. Zhao et al., 2024) | - Expanding research to domains like medical imaging, ocean surveillance, and precision agriculture.<br>- Incorporating SOD models into intelligent systems like SLAM for improved localization accuracy. |
|---|---|
| Real-Time Performance Optimization (Jiang et al., 2024; J. Ni et al., 2024; H. Wang et al., 2024; F. Zhao et al., 2024) | - Enhancing real-time performance by minimizing overhead via pruning, quantization, and architectural refinements.<br>- Striving for lightweight deployment on resource-constrained devices such as UAVs and IoT sensors. |

## 8- Conclusion

SOD remains a vital yet challenging task in computer vision, with significant implications across domains such as surveillance, autonomous systems, medical imaging, and remote sensing. The unique difficulties posed by small objects—such as their limited resolution, susceptibility to background interference, and class imbalance—have driven researchers to innovate with advanced deep learning techniques, including multi-scale feature extraction, attention mechanisms, and SR approaches. The emergence of datasets tailored for SOD, along with robust evaluation metrics, has further advanced this field by enabling more accurate benchmarking and model development. Recent trends, such as the adoption of lightweight architectures, transformer-based models, and KD, have shown great promise in enhancing efficiency and scalability. Additionally, the integration of multi-modal data and domain-specific strategies continues to expand the applications of SOD, addressing real-world challenges in increasingly diverse and complex environments. Looking ahead, addressing unresolved issues like computational efficiency, dataset generalization, and model interpretability will be pivotal in advancing the field. By exploring innovative methodologies and leveraging interdisciplinary insights, SOD is poised to play a transformative role in solving critical problems across industries and ensuring the reliability of intelligent systems in challenging scenarios.

## Statements and Declarations

There are no conflicts of interests to declare.